\pdfoutput=1

\documentclass[11pt]{article}

\usepackage{ACL2023}

\usepackage{times}
\usepackage{latexsym}

\usepackage[T1]{fontenc}

\usepackage[utf8]{inputenc}

\usepackage{microtype}

\usepackage{inconsolata}

\usepackage{comment}
\usepackage{bm}
\usepackage{xcolor}
\usepackage{multirow}
\usepackage{makecell}
\usepackage{textcomp}
\usepackage{graphicx}

\usepackage{makecell}
\usepackage{longtable}
\usepackage{subcaption}
\usepackage{lipsum}
\usepackage{multicol}

\usepackage{amsmath}
\title{INTapt: Information-Theoretic Adversarial Prompt Tuning for Enhanced Non-Native Speech Recognition}

\author{Eunseop Yoon$^{1}$\Thanks{ Equal contribution}, Hee Suk Yoon$^{1}$\footnotemark[1], John Harvill$^{2}$, \\ 
\bf{Mark Hasegawa-Johnson}$^{2}$, \bf{and Chang D. Yoo}$^{1}$\Thanks{ Corresponding author} \\
         $^{1}$Korea Advanced Institute of Science and Technology (KAIST) \\
         $^{2}$University of Illinois at Urbana-Champaign (UIUC) \\ \texttt{\{esyoon97, hskyoon,  cd\_yoo\}@kaist.ac.kr} \\ \texttt{\{harvill2, jhasegaw\}@illinois.edu}}

\begin{document}
\maketitle
\begin{abstract}
Automatic Speech Recognition (ASR) systems have attained unprecedented performance with large speech models pre-trained based on self-supervised speech representation learning. However, these pre-trained speech models suffer from representational bias as they tend to better represent those prominent accents (i.e., native (L1) English accent) in the pre-training speech corpus than less represented accents, resulting in a deteriorated performance for non-native (L2) English accents. Although there have been some approaches to mitigate this issue, all of these methods require updating the pre-trained model weights. In this paper, we propose Information Theoretic Adversarial Prompt Tuning (INTapt), which introduces prompts concatenated to the original input that can re-modulate the attention of the pre-trained model such that the corresponding input resembles a native (L1) English speech without updating the backbone weights. INTapt is trained simultaneously in the following two manners: (1) adversarial training to reduce accent feature dependence between the original input and the prompt-concatenated input and (2) training to minimize CTC loss for improving ASR performance to a prompt-concatenated input. Experimental results show that INTapt improves the performance of L2 English and increases feature similarity between L2 and L1 accents.

\end{abstract}

\section{Introduction}

Self-supervised learning has improved input data representation without requiring extensive human-labeled data. Powerful pre-trained models providing high-performing representations for various data types (e.g., text, images, and audio) have been proposed. For instance, in speech, self-supervised pre-trained models such as HuBERT \citep{HuBERT} have advanced state-of-the-art performance of automatic speech recognition (ASR).




However, one major challenge in using pre-trained speech models for ASR is the representational bias towards prominent accents present in the dataset during pre-training. Consequently, there will be a disparity in ASR performance between native and non-native speakers. More specifically, pre-training using a large dataset such as the LibriSpeech \citep{libri}, which comprises a large proportion of utterances from native (L1) English speakers, leads to a less satisfactory recognition rate for non-native (L2) English accented speech.
This phenomenon can curtail the effectiveness of current high-performing ASR systems for real-world applications.


\begin{figure}[t]
	\centering
	\includegraphics[width=\linewidth]{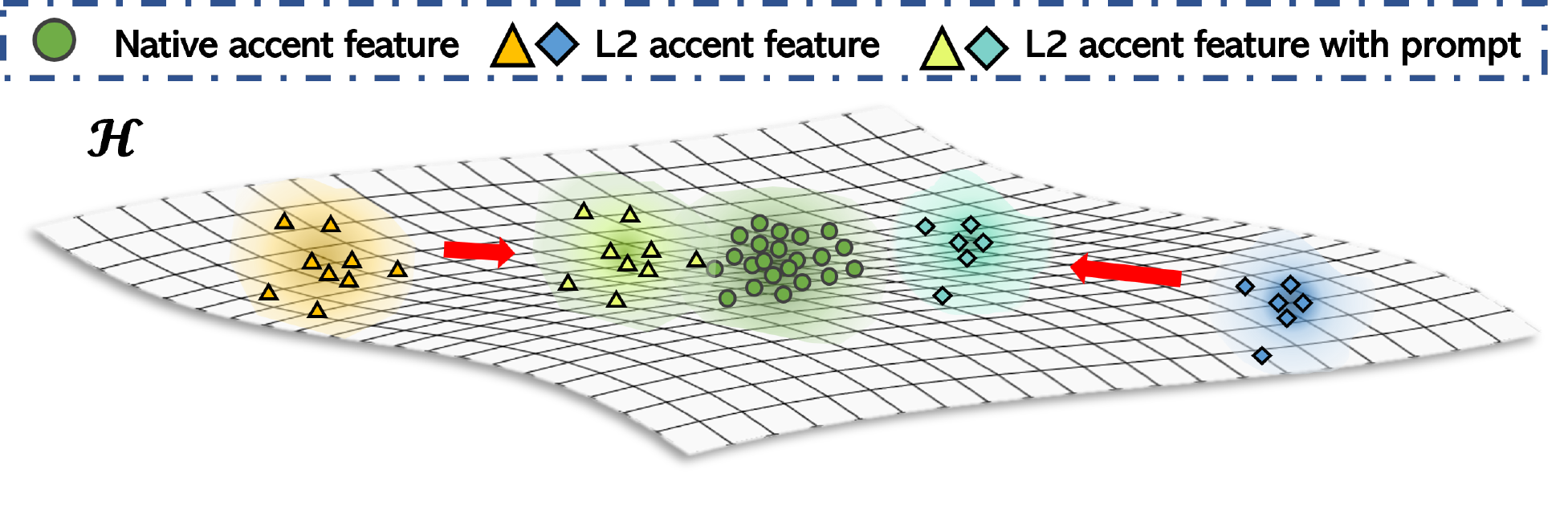}
	\caption{Illustration of a hypothetical accent feature space ($\mathcal{H}$). Distinctive accent features between native and L2 accents lead to degraded performance of ASR systems on L2 accents. INTapt concatenates a prompt to the input space to reduce this distinction.}
	\label{fig:1}
\end{figure}
\begin{figure*}[t]
	\centering
	\includegraphics[width=0.95\linewidth]{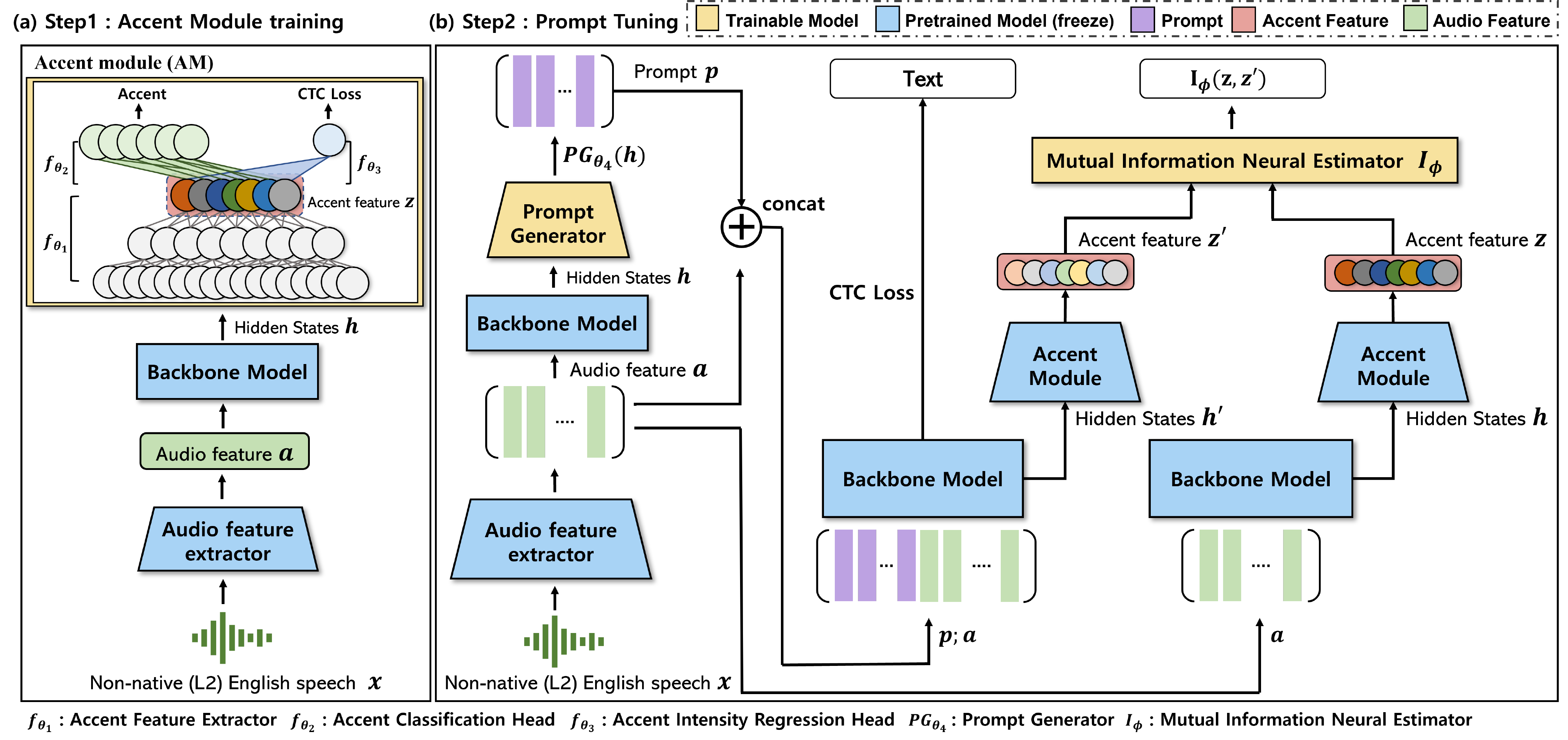}
	\caption{Overview of INTapt. INTapt incorporates a two-step training process where the first step involves training the Accent Module to get the accent feature of a particular input speech and the second step involves training the Prompt Generator capable of making the non-native (L2) English speech input have a better ASR performance by re-modulating the attention of the Backbone Model so that it resembles the accent of a native (L1) English speech.} 
	\label{fig:2}
\end{figure*}

There have been several ways to address this issue in ASR, including fine-tuning the model on diverse accents \citep{prev2, l2arctic_finetune}, having a separate model for each accent \citep{prev1} or using regularization losses that guide the fine-tuning process to achieve robustness to accents \citep{prev3}, all of which require updating the pre-trained model weights.

We propose a different solution for improving L2 speech recognition in transformer-based speech models that introduces a small number of learnable parameters into the input space while keeping the backbone weights of the model untouched. Our approach is guided by Information-Theoretic Adversarial Learning; thus, we refer to it as INTapt (Information-Theoretic Adversarial Prompt Tuning). INTapt aims to introduce auxiliary embeddings (i.e., prompt) concatenated to the original input, which can re-modulate the attention and adapt the pre-trained weights so that the corresponding input looks like speech with an accent seen during pre-training (Figure \ref{fig:1}). To achieve this, INTapt incorporates (1) adversarial training, which tries to minimize the mutual information between the accent feature of the original input and that obtained by concatenating the prompt embeddings in front of the initial input, and (2) CTC loss training to improve the ASR performance of the prompt-concatenated input. Essentially the prompt is trained such that the accent of the concatenation is pushed away from the input accent and the concatenation achieves native CTC loss performance.
Unlike the previous use-case of prompts in NLP or Computer vision (CV), where a single prompt embedding is learned for each discrete task or input domain, the intensity of an accent is continuous. Thus, we propose an input-dependent prompt embedding by training a prompt generator that outputs an input-specific prompt. Through extensive experiments, we show that the proposed dual objectives of INTapt not only lead to better performance on L2 English accents but result in a higher similarity between the accent feature of the prompt-concatenated input and that of L1 English accents.



\section{INTapt}
Figure \ref{fig:2} depicts the overall process of INTapt. INTapt incorporates a two-step training process. In the first step, we train an Accent Module (AM) capable of isolating the accent feature from a given audio feature $\bm{a}$ of an input speech $\bm{x}$. In the second step, we train a Prompt Generator (PG), which outputs a prompt $\bm{p}$ for a given audio feature $\bm{a}$, using two objectives: (1) Minimize the mutual information between the accent feature $\bm{z'}$ and $\bm{z}$, where the former is obtained using the prompt-concatenated input $(\bm{p};\bm{a})$ and the latter is obtained from the original audio feature $\bm{a}$, (2) Minimize CTC loss to improve the ASR performance of the input $(\bm{p};\bm{a}).$ 
\subsection{Accent Module (AM)}

Since our method requires direct access to the isolated accent feature of the corresponding audio feature input, we propose an Accent Module (AM) capable of extracting the accent feature $\bm{z}$ from the input $\bm{a}$. The module consists of an accent feature extractor $f_{\theta_{1}}$ which is trained with an accent classification head $f_{\theta_{2}}$ to isolate the accent feature and an accent intensity regression head $f_{\theta_{3}}$ to capture the intensity of the accent into the obtained feature.

\paragraph{Accent Classification Head} 
The role of the accent classification head $f_{\theta_{2}}$ is to isolate the accent feature of a given speech \footnote{We show in Appendix \ref{accent iso} that the proposed way effectively isolate the accent feature from other features.}. Given the hidden state representation $\bm{h}$ of an audio feature input $\bm{a}$, the feature extractor outputs the accent feature (i.e., $\bm{z} = f_{\theta_{1}}(\bm{h}))$ and the accent classification head $f_{\theta_{2}}$ tries to assign it to the correct accent label $y$.

\paragraph{Accent Intensity Regression Head} 
The intensity of an accent could vary among different people even though there are in the same L2 group, and it could also vary between utterances from the same speaker. Thus, an accent intensity regression head is introduced to incorporate the accent intensity into the obtained accent feature $\bm{z}$.  Based on the assumption that the intensity of the accent affects ASR performance, making the accent intensity regression head predict the CTC loss \footnote{Connectionist Temporal Classification (CTC) \citep{ctc} is the primary loss used to train deep neural networks in speech recognition.}, obtained by inputting the corresponding speech into the backbone speech model, will allow the extracted accent feature $\bm{z}$ to capture the intensity of the accent. 

Given a batch $B$, the training of the Accent Module with the two aforementioned heads could be summarized as:
\begin{equation}
\begin{split}
&\min_{\theta_{1},\theta_{2}} \frac{1}{|B|}\sum_{i\in B}-\log p(y_i|f_{\theta_2}(f_{\theta_1}(h_i))) + \\
&\lambda\min_{\theta_{1},\theta_{3}} \frac{1}{|B|}\sum_{i\in B}	[ \,f_{\theta_{3}}(f_{\theta_{1}}(h_i))-\text{CTC}(x_i)] \,^2 \;\;
\end{split}
\label{eq:1}
\end{equation}


\subsection{Prompt Generator (PG)} 
Building on the success of prompts in NLP \citep{nlp_prompt1, nlp_prompt2} and CV \citep{vit}, we introduce a prompt tuning method to improve the ASR performance for L2 English speech by efficiently utilizing a pre-trained model that already shows good performance for L1 English speech. In contrast to traditional NLP or CV applications, where a single, discrete prompt embedding is learned for each specific task or input domain, the intensity of an accent is continuous. To address this, we propose an input-dependent prompt embedding by training prompt generator $PG_{\theta_{4}}$ that generates an input-specific prompt guided by Information-Theoretic Adversarial Learning. More specifically, given 
a hidden state $\bm{h} = [h_1, h_2, ..., h_L]$ with length $L$ we produce a prompt of length $L'$, 
\begin{equation}
\bm{p} = PG_{\theta_4}(\bm{h})\;\;
\label{eq:2}
\end{equation}

\paragraph{Mutual Information Minimization} 
Mutual Information meausures the co-dependence between two random variables $X$ and $Y$.
\citet{mine} recently proposed a gradient descent based method for estimating this property, allowing the use of neural networks for the estimation of mutual information between high dimensional random variables. The estimation is done using a neural network parameterized by $\phi$ as below:
\begin{equation}
I(X,Y) \geq I_{\phi}(X,Y),
\label{eq:3}
 \end{equation}
where maximizing $I_{\phi}(X,Y)$ provides a tight lower bound of the original mutual information $I(X,Y).$ We use this to adversarially train the prompt generator $PG_{\theta_{4}}$ to minimize the mutual information between the accent feature of the original L2 speech input and the prompt-concatenated input. 

\paragraph{CTC Loss Minimization} We train the prompt generator $PG_{\theta_{4}}$ to minimize the CTC loss obtained for the prompt-concatenated input $(\bm{p}; \bm{a})$.

The two minimization objectives wrt. the prompt generator, along with the maximization objective wrt. the Mutual Information Neural Estimator, are done jointly in the second training step (Equation \ref{step2}). We show in Section \ref{result} and \ref{ablation} that the aforementioned objectives not only improve the ASR performance of L2 speech but also effectively make it resemble the accent feature of the L1 speech.
\begin{equation}
\min_{\theta_{4}}\max_{\phi}\frac{1}{|B|}\sum_{i\in B}\text{CTC}(\bm{p}_{\theta_{4}};\bm{a}) + \lambda I_{\phi}(\bm{z'}_{\theta_{4}},\bm{z})\;
\label{step2}
\end{equation}
\section{Experiments}
\subsection{Experimental setting}
\paragraph{Dataset} We use the L2-ARCTIC \citep{L2arctic}, which is a speech corpus of non-native (L2) English speakers - Mandarin (ZH), Hindi (HI), Vietnamese (VI), Korean (KO), Spanish (ES), and Arabic (AR). Each L2 group contains two male and two female speakers, and all the speakers read the same 1132 texts. The train/dev/test set is configured by dividing the data into 0.8/0.1/0.1 splits with no overlapping texts between each splits. Additionally, since we would like to simulate a natural data collection situation where the amount of data varies across groups, we randomly divided the training data into More Frequent Accent (MFA) (ZH, HI), Less Frequent Accent (LFA) (VI, KO), and Unseen Accent (UA) (ES, AR) - For MFA we keep all the training data, for LFA we keep half of the data, and for UA we remove all the training data. 

\begin{table*}[t]
	\centering
	\begin{tabular}{l||c||l||c c c c c c c}
		\Xhline{2\arrayrulewidth}
		\multirow{2}{*}{Backbone}    &\multirow{2}{*}{ \#.params}     &\multirow{2}{*}{Methods}     & \multicolumn{2}{c}{MFA}     & \multicolumn{2}{c}{LFA}  &\multicolumn{2}{c}{UA}          &\multirow{2}{*}{ALL} \\  
		                                   & & &\begin{small}ZH\end{small} &\begin{small}HI\end{small}&\begin{small}VI\end{small} &\begin{small}KO\end{small}&\begin{small}ES\end{small} &\begin{small}AR\end{small}\\
		\Xhline{2\arrayrulewidth}
    		\multirow{4}{*}{\begin{small}HUBERT
    		$_{Large}$\end{small}}& -& \begin{small}Backbone\end{small}     & \begin{small}18.71\end{small} & \begin{small}8.80\end{small}  & \begin{small}25.8\end{small}  & \begin{small}10.98\end{small}  & \begin{small}14.12\end{small}  & \begin{small}14.92\end{small}      & \begin{small}15.55\end{small}\\
		&315M&\begin{small} +Finetune\end{small}   & \begin{small}15.46\end{small}  & \begin{small}7.91\end{small}  & \begin{small}22.26\end{small}  & \begin{small}9.95\end{small}  & \begin{small}14.19\end{small}  & \begin{small}13.94\end{small}    & \begin{small}13.95\end{small}\\ 
		&12.5M&\begin{small} +Prompt$_{ctc}$\end{small} & \begin{small}13.93\end{small}  & \begin{small}7.20\end{small}  & \begin{small}21.93\end{small}   & \begin{small}9.69\end{small}  & \begin{small}12.64\end{small} & \begin{small}12.38\end{small}   & \begin{small}12.96\end{small}\\
		&12.9M&\begin{small} +INTapt\end{small} &\begin{small}\bf{13.09}\end{small}  &\begin{small} \bf{6.64}\end{small}  & \begin{small}\bf{21.25}\end{small}  &\begin{small} \bf{8.97}\end{small}  & \begin{small}\bf{12.18}\end{small}  & \begin{small}\bf{11.92}\end{small}  & \begin{small}\bf{12.34}\end{small}
		\\\hline
		\multirow{4}{*}{\begin{small}HUBERT$_{XLarge}$\end{small}} &-& \begin{small}Backbone\end{small}     & \begin{small}17.03\end{small} & \begin{small}7.48\end{small}  & \begin{small}26.02\end{small}  & \begin{small}10.49\end{small}  & \begin{small}13.65\end{small}  & \begin{small}13.52\end{small}    & \begin{small}14.69\end{small} \\
		&958M&\begin{small} +Finetune\end{small}   & \begin{small}15.49\end{small}  & \begin{small}7.53\end{small}  & \begin{small}24.09\end{small}  & \begin{small}10.02\end{small}  & \begin{small}13.48\end{small}  & \begin{small}12.56\end{small}   & \begin{small}13.86\end{small}\\ 
		&19.7M&\begin{small} +Prompt$_{ctc}$\end{small} & \begin{small}13.02\end{small}  & \begin{small}7.31\end{small}  & \begin{small}19.26\end{small}   & \begin{small}8.05\end{small}  & \begin{small}10.46\end{small} & \begin{small}\bf{10.38}\end{small}  & \begin{small}11.41\end{small}\\
		&19.9M&\begin{small} +INTapt\end{small} &\begin{small}\bf{11.67}\end{small}  &\begin{small}\bf{6.63}\end{small}  & \begin{small}\bf{18.41}\end{small}  &\begin{small} \bf{7.17}\end{small}  & \begin{small}\bf{10.44}\end{small}  & \begin{small}10.55\end{small} & \begin{small}\bf{11.00}\end{small}
		\\\hline

		\Xhline{3\arrayrulewidth} 

	\end{tabular}
	\caption{Comparison of WER (\%) (lower is better) on the created subset of L2-ARCTIC (MFA, LFA, UA). \#.params denote the number of parameters that were updated for training. The std. are reported in Table \ref{tab:table_Std} of Appendix \ref{sec:appendix_std}.} 
	\label{tab:table1}
\end{table*}

\paragraph{Models} For the backbone pre-trained speech models we try two different settings, HuBERT$_{Large}$ and HuBERT$_{XLarge}$ \cite{HuBERT}.  We consider three different training situations: 1) \textbf{Finetune} denotes a standard finetuning method where we update the pre-trained model weights to minimize the CTC loss, 2) \textbf{Prompt}$_{ctc}$ is the case of training the prompt generator without the minimization of mutual information, and 3) \textbf{INTapt} trains the prompt generator with our proposed objective in equation \ref{step2}. We include the training details in Appendix \ref{sec:appendidx_training}.

\subsection{Results}
\label{result} 
Table \ref{tab:table1} shows the Word Error Rate (WER) across different L2 groups on the ASR task. We find that the performance improvement of the prompt tuning approaches (Prompt$_{ctc}$ and INTapt) are more significant compared to standard finetuning despite updating small number of parameters (2-4\%). INTapt shows the lowest WER on all L2 groups, obtaining 12.34\% for HuBERT$_{Large}$  and 11.00\% for HuBERT$_{XLarge}$  on the aggregated all speakers, outperforming the finetuned by 1.62\%p and 2.86\%p, respectively \footnote{We show some examples of improved ASR results using INTapt in Appendix \ref{example}}. This conforms to the previous findings \citep{prompt_modelscale} that larger model size can benefit more from prompt tuning methods. 

In Table \ref{tab:libri}, we report the WER on LibriSpeech \citep{libri} test-clean and test-other, which consists mainly of L1 speech. Compared with the backbone model, the WER after finetuning increased by 5.81\%p. However, since Prompt$_{ctc}$ and INTapt does not change the backbone weights, the WER on test-all increased only by 0.48\%p and 0.37\%p, respectively. This shows one of the key benefits of prompt tuning methods in that it only slightly degrades the performance of the backbone model on tasks it already excels at while improving performance on others.

\begin{table}[t]
	\centering
	\begin{tabular}{l||c c c }
		\Xhline{2\arrayrulewidth}
        Methods    & test-clean     & test-other  &test-all\\   
		                                   
		\Xhline{2\arrayrulewidth}
	     \begin{small}Backbone\end{small}     & 2.15 & 4.42  & 3.29  \\
		\begin{small} +Finetune\end{small}    & 8.10 & 10.08  & 9.10   \\ 
		\begin{small} +Prompt$_{ctc}$\end{small}  & 2.56 & 4.93  & 3.77    \\
		\begin{small} +INTapt\end{small} & 2.41 & 4.94  & 3.66 
		\\\hline
	\end{tabular}
	\caption{WER (\%) (lower is better) on LibriSpeech. test-all denotes the aggregation of test-clean and test-other.}
	\label{tab:libri}
\end{table}

\begin{figure}[h]
	\centering
	\includegraphics[width=0.7\linewidth]{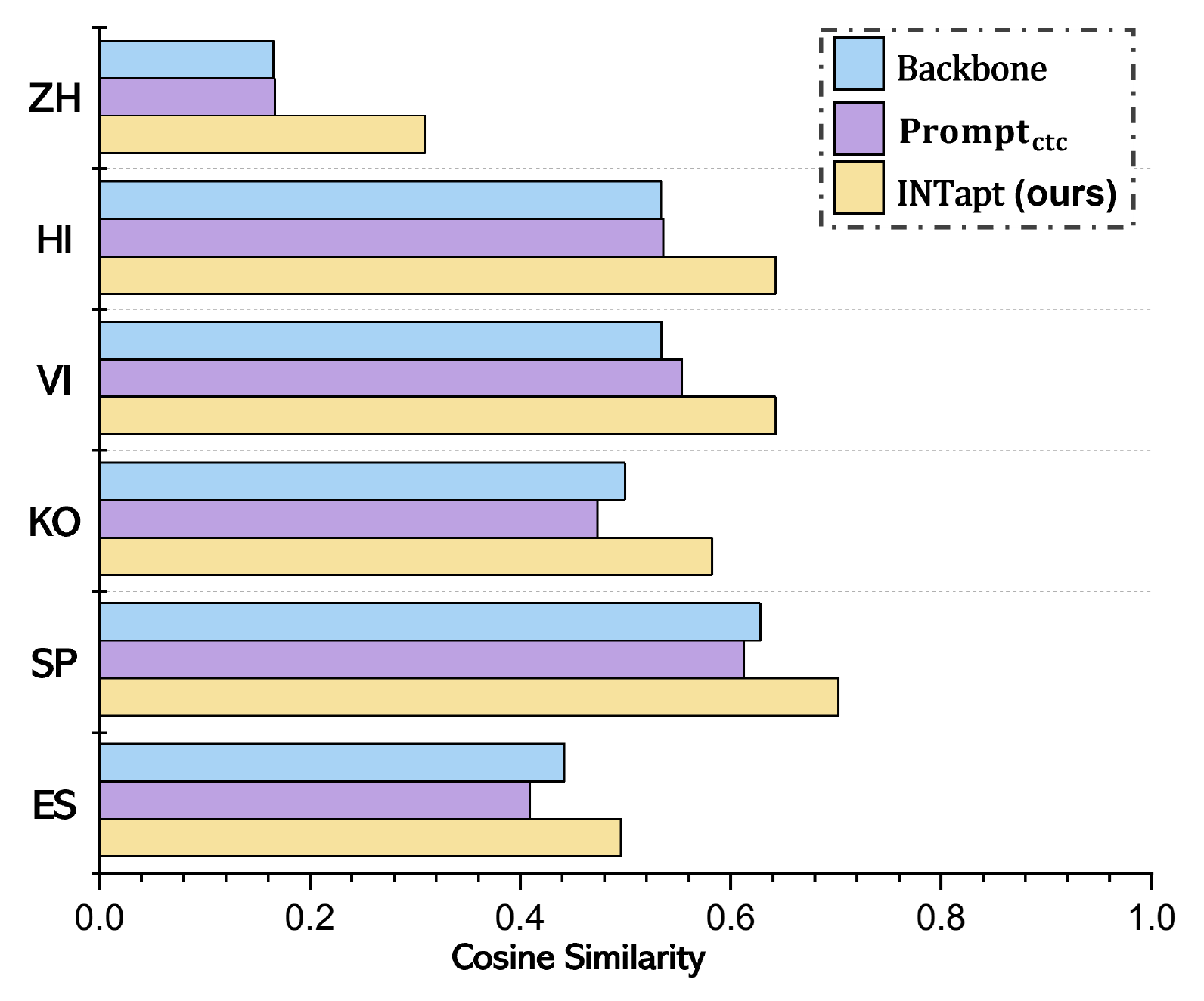}
	\caption{Cosine similarity between L1 accent feature and L2 accent features obtained from different methods.}
	\label{fig:3}
\end{figure}

\section{Ablation Study}
\label{ablation}

We analyze whether INTapt allows the L2 speech input to resemble the accent of L1 speech (Figure \ref{fig:3}). Using the Accent Module, we extract the L1 accent feature, and L2 accent features obtained using the Backbone model, Prompt$_{ctc}$, and INTapt. INTapt showed the highest cosine similarity for all L2 groups, meaning that INTapt effectively adjusts the attention of the pre-trained model so that L2 speech resemble L1 speech in terms of accent. 
 
\section{Conclusion}
We introduced Information Theoretic Adversarial Prompt Tuning (INTapt) for improving non-native ASR performance. To achieve this, INTapt re-modulates the attention of the pre-trained speech models by concatenating input-dependent prompt embeddings to the original input, without updating the model weights. Throughout the experiment, we show that INTapt is capable of outperforming standard finetuning of the pre-trained model on L2 speech, without degradation on L1 speech, by allowing the L2 input to resemble a L1 accent.


\section*{Limitations}
INTapt adopts a prompt tuning method which utilizes the inherent information from pre-trained models that already shows good ASR performance on L1 English speakers. 
Therefore, in order to apply our method we need a pre-trained model that already has good performance on a specific task which might not be available for other langauges.
Also, our method might potentially need sufficiently large pre-trained model size in order for prompt to utilize the internal information of the model.

\section*{Ethics Statement}
Since pre-trained speech models usually show better performance on native (L1) speech automatic speech recognition (ASR) due to the nature of the pre-training data used, this work  have contributed to improve the ASR performance for non-native (L2) English speakers and mitigating the performance gap between them. This has the potential to construct a fair ASR machine well-operating not only on L1 English speakers but L2 speakers, which is an important feature to have for its deployment in real-life. Additionally, since we utilize the pre-trained model, it is possible to have ethical issue depending on the pre-trained model.  

\section*{Acknowledgements}
This work was supported by Institute of Information \& communications Technology Planning \& Evaluation (IITP) grant funded by the Korea government (MSIT) (No. 2022-0-00951, Development of Uncertainty-Aware Agents Learning by Asking Questions), Institute of Information \& communications Technology Planning \& Evaluation (IITP) grant funded by the Korea government(MSIT) [NO.2021-0-02068, Artificial Intelligence Innovation Hub (Seoul National University)], and Institute of Information \& communications Technology Planning \& Evaluation (IITP) grant funded by the Korea government(MSIT) (No.2022-0-00184, Development and Study of AI Technologies to Inexpensively Conform to Evolving Policy on Ethics).

\bibliography{anthology,custom}

\begin{thebibliography}{15}
\expandafter\ifx\csname natexlab\endcsname\relax\def\natexlab#1{#1}\fi

\bibitem[{Belghazi et~al.(2018)Belghazi, Baratin, Rajeswar, Ozair, Bengio,
  Courville, and Hjelm}]{mine}
Mohamed~Ishmael Belghazi, Aristide Baratin, Sai Rajeswar, Sherjil Ozair, Yoshua
  Bengio, Aaron Courville, and R~Devon Hjelm. 2018.
\newblock \href {https://doi.org/10.48550/ARXIV.1801.04062} {Mine: Mutual
  information neural estimation}.

\bibitem[{Chen et~al.(2020)Chen, Yang, Yeh, Jain, and Seltzer}]{prev3}
Yi-Chen Chen, Zhaojun Yang, Ching-Feng Yeh, Mahaveer Jain, and Michael~L
  Seltzer. 2020.
\newblock Aipnet: Generative adversarial pre-training of accent-invariant
  networks for end-to-end speech recognition.
\newblock In \emph{ICASSP 2020-2020 IEEE International Conference on Acoustics,
  Speech and Signal Processing (ICASSP)}, pages 6979--6983. IEEE.

\bibitem[{Dosovitskiy et~al.()Dosovitskiy, Beyer, Kolesnikov, Weissenborn
  et~al.}]{vit}
A~Dosovitskiy, L~Beyer, A~Kolesnikov, D~Weissenborn, et~al.
\newblock An image is worth 16x16 words: Transformers for image recognition at
  scale.

\bibitem[{Graves et~al.(2006)Graves, Fern\'{a}ndez, Gomez, and
  Schmidhuber}]{ctc}
Alex Graves, Santiago Fern\'{a}ndez, Faustino Gomez, and J\"{u}rgen
  Schmidhuber. 2006.
\newblock \href {https://doi.org/10.1145/1143844.1143891} {Connectionist
  temporal classification: Labelling unsegmented sequence data with recurrent
  neural networks}.
\newblock In \emph{Proceedings of the 23rd International Conference on Machine
  Learning}, ICML '06, page 369–376, New York, NY, USA. Association for
  Computing Machinery.

\bibitem[{Hsu et~al.(2021)Hsu, Bolte, Tsai, Lakhotia, Salakhutdinov, and
  Mohamed}]{HuBERT}
Wei-Ning Hsu, Benjamin Bolte, Yao-Hung~Hubert Tsai, Kushal Lakhotia, Ruslan
  Salakhutdinov, and Abdelrahman Mohamed. 2021.
\newblock Hubert: Self-supervised speech representation learning by masked
  prediction of hidden units.
\newblock \emph{IEEE/ACM Transactions on Audio, Speech, and Language
  Processing}, 29:3451--3460.

\bibitem[{Lester et~al.(2021)Lester, Al-Rfou, and Constant}]{prompt_modelscale}
Brian Lester, Rami Al-Rfou, and Noah Constant. 2021.
\newblock \href {https://doi.org/10.18653/v1/2021.emnlp-main.243} {The power of
  scale for parameter-efficient prompt tuning}.
\newblock In \emph{Proceedings of the 2021 Conference on Empirical Methods in
  Natural Language Processing}, pages 3045--3059, Online and Punta Cana,
  Dominican Republic. Association for Computational Linguistics.

\bibitem[{Li and Liang(2021)}]{nlp_prompt2}
Xiang~Lisa Li and Percy Liang. 2021.
\newblock Prefix-tuning: Optimizing continuous prompts for generation.
\newblock In \emph{Proceedings of the 59th Annual Meeting of the Association
  for Computational Linguistics and the 11th International Joint Conference on
  Natural Language Processing (Volume 1: Long Papers)}, pages 4582--4597.

\bibitem[{Liu et~al.(2021)Liu, Yuan, Fu, Jiang, Hayashi, and
  Neubig}]{nlp_prompt1}
Pengfei Liu, Weizhe Yuan, Jinlan Fu, Zhengbao Jiang, Hiroaki Hayashi, and
  Graham Neubig. 2021.
\newblock Pre-train, prompt, and predict: A systematic survey of prompting
  methods in natural language processing.
\newblock \emph{arXiv preprint arXiv:2107.13586}.

\bibitem[{Loshchilov and Hutter(2019)}]{Adam}
Ilya Loshchilov and Frank Hutter. 2019.
\newblock \href {https://openreview.net/forum?id=Bkg6RiCqY7} {Decoupled weight
  decay regularization}.
\newblock In \emph{International Conference on Learning Representations}.

\bibitem[{Panayotov et~al.(2015)Panayotov, Chen, Povey, and Khudanpur}]{libri}
Vassil Panayotov, Guoguo Chen, Daniel Povey, and Sanjeev Khudanpur. 2015.
\newblock \href {https://doi.org/10.1109/ICASSP.2015.7178964} {Librispeech: An
  asr corpus based on public domain audio books}.
\newblock In \emph{2015 IEEE International Conference on Acoustics, Speech and
  Signal Processing (ICASSP)}, pages 5206--5210.

\bibitem[{Shibano et~al.(2021)Shibano, Zhang, Li, Cho, Sullivan, and
  Abdul-Mageed}]{l2arctic_finetune}
Toshiko Shibano, Xinyi Zhang, Mia~Taige Li, Haejin Cho, Peter Sullivan, and
  Muhammad Abdul-Mageed. 2021.
\newblock Speech technology for everyone: Automatic speech recognition for
  non-native english.
\newblock In \emph{Proceedings of The Fourth International Conference on
  Natural Language and Speech Processing (ICNLSP 2021)}, pages 11--20.

\bibitem[{van~der Maaten and Hinton(2008)}]{tsne}
Laurens van~der Maaten and Geoffrey Hinton. 2008.
\newblock \href {http://jmlr.org/papers/v9/vandermaaten08a.html} {Visualizing
  data using t-sne}.
\newblock \emph{Journal of Machine Learning Research}, 9(86):2579--2605.

\bibitem[{Winata et~al.(2019)Winata, Lin, and Fung}]{prev2}
Genta~Indra Winata, Zhaojiang Lin, and Pascale Fung. 2019.
\newblock \href {https://doi.org/10.18653/v1/W19-4320} {Learning multilingual
  meta-embeddings for code-switching named entity recognition}.
\newblock In \emph{Proceedings of the 4th Workshop on Representation Learning
  for NLP (RepL4NLP-2019)}, pages 181--186, Florence, Italy. Association for
  Computational Linguistics.

\bibitem[{Yang et~al.(2018)Yang, Audhkhasi, Rosenberg, Thomas, Ramabhadran, and
  Hasegawa-Johnson}]{prev1}
Xuesong Yang, Kartik Audhkhasi, Andrew Rosenberg, Samuel Thomas, Bhuvana
  Ramabhadran, and Mark Hasegawa-Johnson. 2018.
\newblock Joint modeling of accents and acoustics for multi-accent speech
  recognition.
\newblock In \emph{2018 IEEE International Conference on Acoustics, Speech and
  Signal Processing (ICASSP)}, pages 1--5. IEEE.

\bibitem[{Zhao et~al.(2018)Zhao, Sonsaat, Silpachai, Lucic,
  Chukharev-Hudilainen, Levis, and Gutierrez-Osuna}]{L2arctic}
Guanlong Zhao, Sinem Sonsaat, Alif Silpachai, Ivana Lucic, Evgeny
  Chukharev-Hudilainen, John Levis, and Ricardo Gutierrez-Osuna. 2018.
\newblock L2-arctic: A non-native english speech corpus.
\newblock In \emph{INTERSPEECH}, pages 2783--2787.

\end{thebibliography}
\bibliographystyle{acl_natbib} 

\newpage
\appendix
\onecolumn
\section{Experiment Details}
\label{sec:appendidx_training}

\paragraph{Model Architecture} We use a 3-layer Multi Layer Perceptron (MLP) for the Accent Feature Extractor and 1-layer, 3-layer MLP for the Accent Classification Head and the Accent Intensity Regression Head in the Accent Module, respectively. The Prompt Generator (PG) is composed of a single layer transformer. Since we adopt the transformer architecture for PG, its maximum output length is same as the length $L$ of the input audio feature $a$. The specific length of prompt can be set by taking the first $L'$ output embeddings from the front of the transformer output. For the Mutual Information Neural Estimator (MINE), we use a 3-layer MLP as well. 


\paragraph{Training Details} For pre-processing L2-ARCTIC, we utilized the huggingface resampling tool \footnote{\url{https://huggingface.co/docs/datasets/audio_process}} to downsample the audio files from 44.1kHz to 16kHz. The hidden state representation obtained from the 3rd layer of the backbone model is used as the input to the Accent Module and Prompt Generator for both HuBERT$_{large}$ and HuBERT$_{XLarge}$. The dimension of the accent feature $\bm{a}$ is set to $d=256$, the length of prompt $L'$ is 40, and the dimension of the prompt is set to 1024 and 1280, same as that of the input embedding size for HuBERT$_{large}$ and HuBERT$_{XLarge}$, respectively. We use the AdamW optimizer \citep{Adam} with $\beta_1 = 0.9, \beta_2 = 0.999, \epsilon=1e-8$, and weight decay $\lambda = 0.005$ with different learning rates for all trainable model (i.e., AM, PG, MINE, finetuned backbone). The learning rate used for both AM and MINE is 1e-3, and 5e-6, 1e-4, 1e-4 are used for Fintune, prompt$_{ctc}$, and INTapt, respectively. For all the methods, the batch size is set to 16 for HuBERT$_{large}$ and 8 for HuBERT$_{XLarge}$. We use $\lambda=0.5$ for Equation \ref{eq:1} and $\lambda=0.003$ for Equation \ref{step2}. The best model is selected by the lowest WER on the validation set. All experiments was done on NVIDIA Quadro RTX 8000.




\section{Accent Feature Isolation}
\label{accent iso}
We visually analyze the accent feature extracted from AM to validate that the feature does not contain any other information except accent. We plot the 2-D representation of extracted accent feature using t-SNE \citep{tsne} with the label for gender for three L2 groups (i.e., HI, KO, ES). Figure \ref{fig:4} shows that the scatter points are distinctive between L2 groups but difficult to distinguish  gender, which means our AM successfully isolates the accent feature from audio.
\begin{figure}[h]
	\centering
	\includegraphics[width=0.6\linewidth]{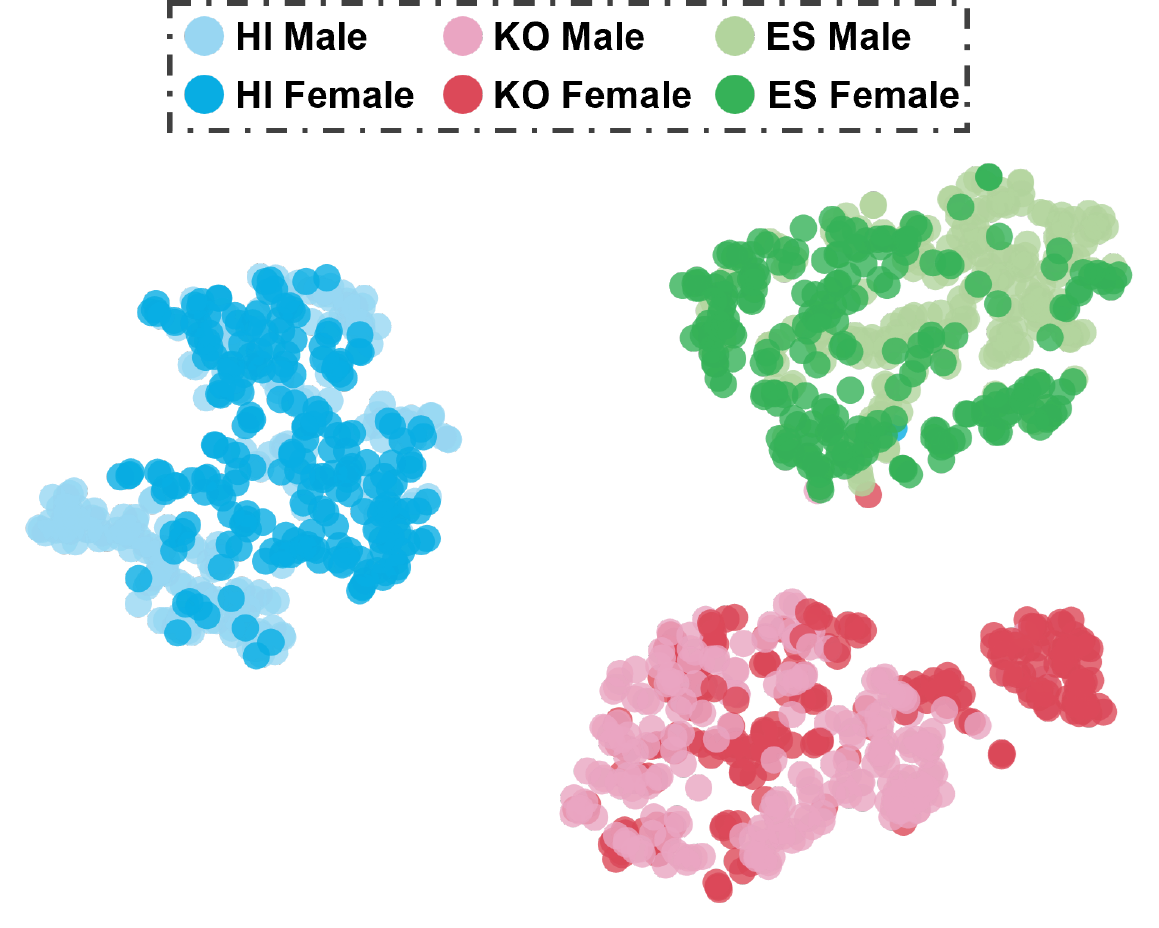}
	\caption{Latent space visualization showing that Accent Module extracts isolated accent feature}
	\label{fig:4}
\end{figure}

\section{Results of INTapt}
\subsection{Examples}
\label{example}
In Figure \ref{fig:4}, we show some examples of improved speech recognition results using INTapt. HuBERT$_{Large}$ is used as backbone model for both cases and the INTapt having lowest WER on val set is selected to predict samples. \textcolor{red}{Red} represents wrong  transcription output from the backbone pre-trained HuBERT$_{Large}$, and \textcolor{green}{green} represents enhanced recognition through the use of INTapt to re-modulate the backbone attention.

\begin{figure*}[h]
	\centering
	\includegraphics[width=\linewidth]{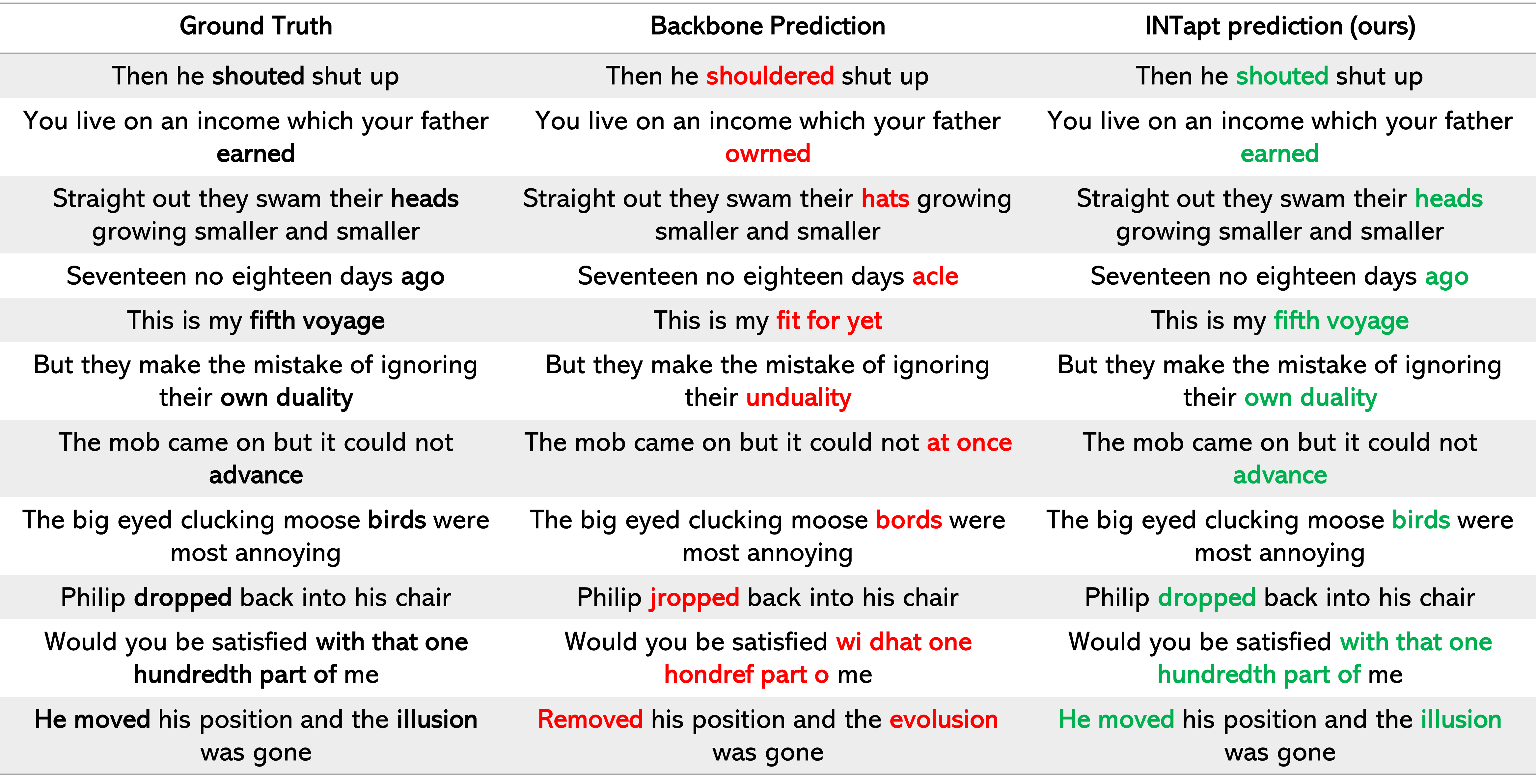}
	\caption{Examples of enhanced recognition though the use of INTapt. \textcolor{red}{Red} represents the wrong transcription output prediction from the backbone model and \textcolor{green}{green} represents the correct output prediction using INTapt.}
	\label{fig:4}
\end{figure*}

\subsection{Standard Deviation of Results}
\label{sec:appendix_std}
\begin{table*}[h]
		\centering
	\begin{tabular}{l||c||l||c c c c c c c}
		\Xhline{2\arrayrulewidth}
		\multirow{2}{*}{Backbone}    &\multirow{2}{*}{ \#.params}     &\multirow{2}{*}{Methods}     & \multicolumn{2}{c}{MFA}     & \multicolumn{2}{c}{LFA}  &\multicolumn{2}{c}{Unseen Accent}          &\multirow{2}{*}{Avg.} \\  
		                                   & & &\begin{small}ZH\end{small} &\begin{small}HI\end{small}&\begin{small}VI\end{small} &\begin{small}KO\end{small}&\begin{small}ES\end{small} &\begin{small}AR\end{small}\\
		\Xhline{2\arrayrulewidth}
		\multirow{3}{*}{\begin{small}HUBERT
    		$_{Large}$\end{small}}&315M&\begin{small} +Finetune\end{small}   & \begin{small}0.31\end{small}  & \begin{small}0.70\end{small}  & \begin{small}0.51\end{small}  & \begin{small}0.50\end{small}  & \begin{small}0.73\end{small}  & \begin{small}0.38\end{small}    & \begin{small}0.36\end{small}\\ 
		&12.5M&\begin{small} +Prompt$_{ctc}$\end{small} & \begin{small}0.51\end{small}  & \begin{small}0.56\end{small}  & \begin{small}0.99\end{small}   & \begin{small}0.74\end{small}  & \begin{small}0.78\end{small} & \begin{small}0.38\end{small}   & \begin{small}0.30\end{small}\\
		&12.9M&\begin{small} +INTapt\end{small} &\begin{small}0.66\end{small}  &\begin{small} 0.72\end{small}  & \begin{small}0.73\end{small}  &\begin{small} 0.69\end{small}  & \begin{small}0.59\end{small}  & \begin{small}0.27\end{small}  & \begin{small}0.13\end{small}
		\\\hline
		\multirow{3}{*}{\begin{small}HUBERT$_{XLarge}$\end{small}}&958M&\begin{small} +Finetune\end{small}   & \begin{small}0.41\end{small}  & \begin{small}0.21\end{small}  & \begin{small}1.79\end{small}  & \begin{small}0.06\end{small}  & \begin{small}0.33\end{small}  & \begin{small}0.12\end{small}   & \begin{small}0.29\end{small}\\ 
		&19.7M&\begin{small} +Prompt$_{ctc}$\end{small} & \begin{small}0.15\end{small}  & \begin{small}0.64\end{small}  & \begin{small}0.19\end{small}   & \begin{small}0.51\end{small}  & \begin{small}0.24\end{small} & \begin{small}0.56\end{small}  & \begin{small}0.12\end{small}\\
		&19.9M&\begin{small} +INTapt\end{small} &\begin{small}0.33\end{small}  &\begin{small}0.31\end{small}  & \begin{small}0.41\end{small}  &\begin{small} 0.30\end{small}  & \begin{small}0.25\end{small}  & \begin{small}0.12\end{small} & \begin{small}0.18\end{small}
		\\\hline

		\Xhline{3\arrayrulewidth} 

	\end{tabular}
	\caption{Standard deviation values for the experimental results in Table \ref{tab:table1}. The values were obtained by running the same experiments with five different random seeds.}
	\label{tab:table_Std}
\end{table*}
In Table \ref{tab:table_Std}, we report the standard deviation of the results in Table \ref{tab:table1} with five different random seeds. As the backbone experiment in Table \ref{tab:table1} is obtained without any training, we do not contain the standard deviation for those.

\end{document}